# IRIS RECOGNITION BASED ON LBP AND COMBINED LVQ CLASSIFIER

M. Z. Rashad[1], M. Y. Shams[2], O. Nomir[2], and R. M. El-Awady[3]

Dept. of Computer Science, Faculty of Computer and Information Sciences, Mansoura University, Egypt.

[1]magdi_Z2011@yahoo.com
[2]{myysd2011, omnomir}@yahoo.com,
[3]ryysd2008@yahoo.com

***ABSTRACT***

*Iris recognition is considered as one of the best biometric methods used for human identification and verification, this is because of its unique features that differ from one person to another, and its importance in the security field. This paper proposes an algorithm for iris recognition and classification using a system based on Local Binary Pattern and histogram properties as a statistical approaches for feature extraction , and Combined Learning Vector Quantization Classifier as Neural Network approach for classification, in order to build a hybrid model depends on both features. The localization and segmentation techniques are presented using both Canny edge detection and Hough Circular Transform in order to isolate an iris from the whole eye image and for noise detection .Feature vectors results from LBP is applied to a Combined LVQ classifier with different classes to determine the minimum acceptable performance, and the result is based on majority voting among several LVQ classifier. Different iris datasets CASIA, MMU1, MMU2, and LEI with different extensions and size are presented. Since LBP is working on a grayscale level so colored iris images should be transformed into a grayscale level. The proposed system gives a high recognition rate 99.87 % on different iris datasets compared with other methods.*

## 1. INTRODUCTION

Every human has a personally identifiable different from others such as shape, size , color of his eyes, voice, and even body odor. Modern sciences employed these differences to distinguish between one person and another, and in almost there is no error. Biometric identification can be classified into two classes; the first class is called *physiological* which interested in the shape of the body like face, fingerprint, hand geometry and iris recognition. The second is called *behavioral* that are related to the behavior of a person like signature and voice [1]. The Iris Recognition System is considered as one of the important ways for security in airports, government buildings, and research laboratories [2]. Iris image contains not only useful parts i.e. iris but also some irrelevant parts i.e. noise like eyelid, pupil, eyelashes, specular highlight. The iris is the annular part between black pupil and white sclera is the most part that researches are focused to determine its details. In general , there are many properties that make an iris ideal biometric method , the first is the uniqueness features " no two iris are the same" even between the left and right eye for the same person [3]. The second is the accuracy results from an iris pattern which is unchanged through a person's life with the data-rich physical structure. Iris recognition system is generally includes a series of steps : (i) image acquisition, (ii) iris preprocessing includes localization, segmentation, and normalization (iii) feature extraction, and

(iv) matching and classification, as shown in figure 1. Image acquisition is the first step in IRS which an iris image is captured, and the second step is preprocessing includes localization, segmentation and normalization. The third step is the feature extraction to get the feature vector and iris signature used for matching and classification to obtain the recognition rate. In this paper both texture analysis and matching of texture representation will be used with the aid of combined classifier learning vector quantization (LVQ) and a comparative evaluation with other methods using different iris datasets will be presented. This paper has been organized as follows (2) Related Work, (3) Preprocessing, (4) Feature extraction, (5) Proposed algorithm, (6) Experimental and discussion, and (7) conclusion.

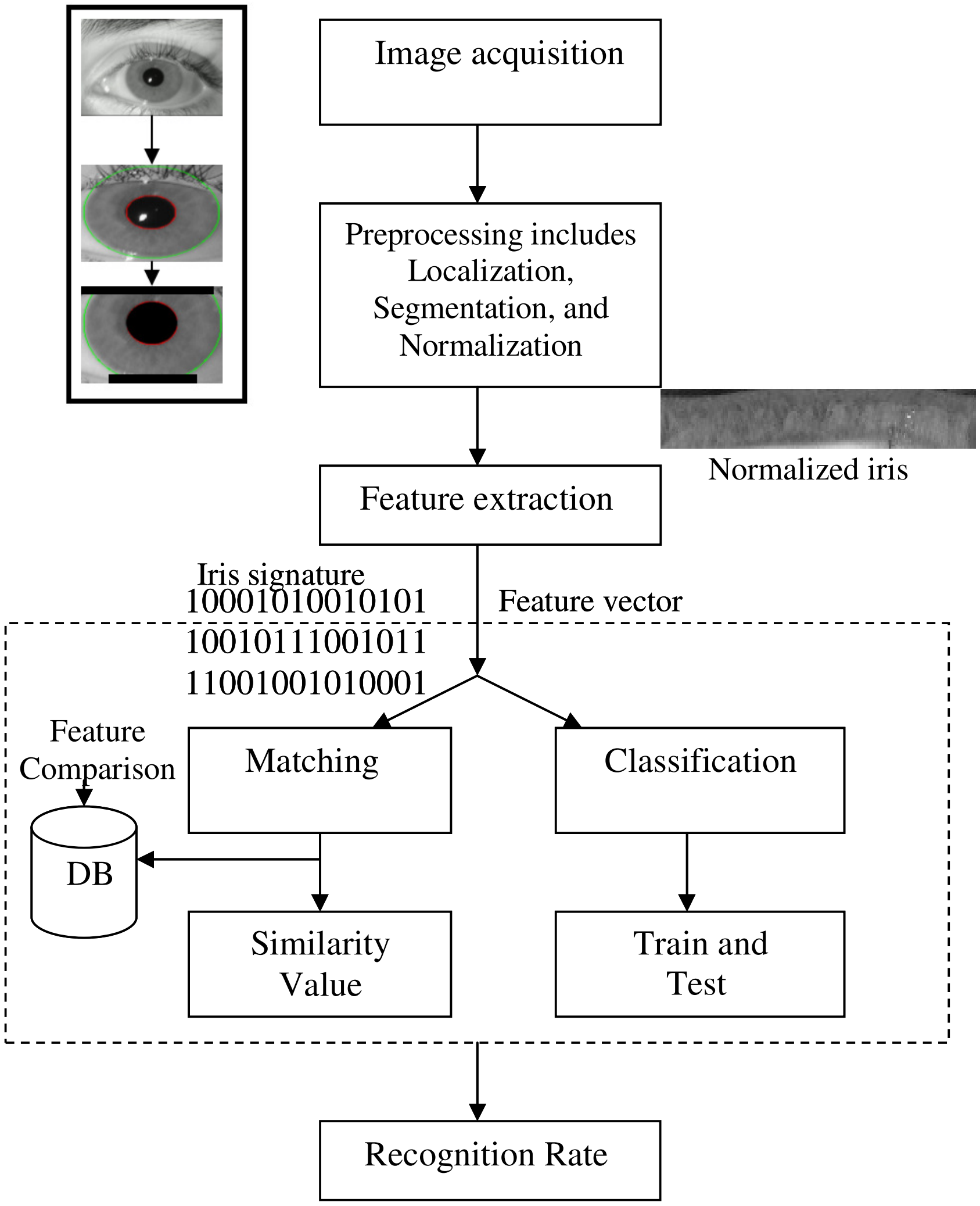

**Fig. 1.** General structure of Iris Recognition System.

## 2. RELATED WORK

A detailed literature survey of iris recognition algorithms can be found in [4]. J. Daugman [5] shows that the failure of a test of statistical independence is the key to iris when any eye's phase code is compared with another version recognition. His test of statistical independence is implemented by XOR applied to the 2048 bit phase vectors. The XOR operator detects the disagreement between any corresponding pair of bits, while the AND operator ∩ ensures that the compared bits are both deemed to have been uncorrupted by eyelashes, eyelids, secular, or other noise. He used Hamming distance (HD) as a measure of the dissimilarity between any two irises. Moreover , J. Daugman [6] presents the following four steps for iris recognition: (i) detecting the iris inner and outer boundaries with active Contours (ii) Fourier-based methods for solving problems in iris trigonometry and projective geometry, (iii) statistical inference methods for detecting and excluding eyelashes, and (iv) exploration of score normalizations.

U. Gawande et al [7], proposes combines of the zero-crossing 1 D wavelet Euler No., and genetic algorithm based for feature extraction. The output from these algorithms is normalized and their score are fused to decide whether the user is genuine or imposter. The performance measure of verification is related to the frequency with which errors happens.

P. Manikandan and M. Sundararajan [8] proposes feature extraction method based on the Discrete Wavelet compared with (2D-DWT) two dimensional-discrete wavelet transform in order to improve the classification accuracy , they used two types of iris databases with the Correct Recognition Rate (CRR) 99.83% and 98.15% on DB1 and DB2 respectively .

T. Karthikeyan [9] proposes an algorithm based on Fuzzy Neural, the located iris after pre-processing is in feature vector used with the neural network to recognize iris patterns. The accuracy of his proposed method for the trained pattern is 99.25%.

L. Fallah et al [10] proposes iris Recognition based on covariance of discrete wavelet using Competitive Neural Network (LVQ) of discrete wavelet .they present a set of edge iris profiles to build a covariance matrix by discrete wavelet transform using Neural Network.

A. Azizi and H. R. Pourreza [11] proposes an algorithm for iris feature extraction using contourlet transform, Contourlet transform captures the intrinsic geometrical structures of iris image using SVM (Support Vector Machine) classifier for approximating people identification to provide a less feature vector length with an insignificant reduction of the percentage of correct classification.

M. Shamsi et al [12] proposes an algorithm that hybrid summation function and factor matrix to be able to detect the iris boundaries, they present difference function and the detection accuracy has been significantly improved, the segmentation rate using this algorithm is 99.34%. In [13] they uses a modify segmentation operator to detect the pupil and iris as ellipse instead of circle because many irises are not exactly circle, and they found that an ellipse view will improve the accuracy of iris segmentation.

## 3. PREPROCESSING

Image preprocessing is a very important step in Iris Recognition System in order to get rid of the image noise, and prepare the iris image to feature extraction with a little noise. The second step after image acquisition is image preprocessing. A standard datasets like MMU1, MMU2 [14], CASIA [15], and Lion's Eye Institute (LEI) [16-17] are presented in this paper as the external input image. Image preprocessing can be divided into, iris localization, iris segmentation and iris normalization.

### 3.1 . Iris localization

Cyrille Baptiste [18] presents an algorithm to localize the eye with a series of steps, by which the using of this algorithm with Circular Hough transform (CHT) and canny edge detection can achieve a good results explained as follows:

**Step 1:** Determination of the mean intensity of an image.
**Step 2:** Determine the Cut off interest part of an image.
**Step 3:** Binarization of the interest part. The threshold is given by the following equation:

$$threshold = 0.52 \left[ \frac{1}{size_{image}} \sum Intensity_{pixels} \right] \qquad (1-a)$$

Binarization of the interest part is shown in figure 2. To specify the unwanted region of an iris image , Canny edge detection is presented with different threshold values between 0.2 to 0.7 as shown in figure 3 .Fig (3-a ) when applying LEI dataset with size 512×256 and JPG format. Fig (3-b) when applying CASIA dataset with size 320×280 and GIF format.

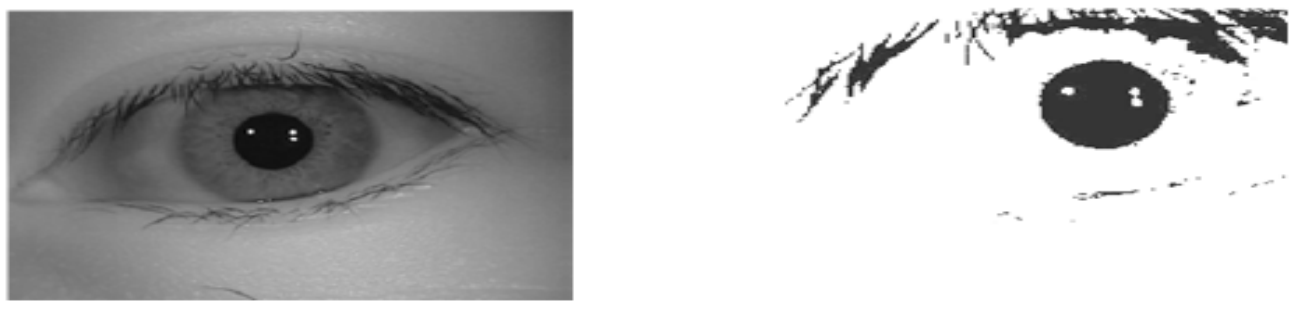

**Fig. 2.** Binarization of the interest part.

**Step 4:** Using HCT (Hough Circle Transform) in order to localize eye image by detecting the projection point of vertical and horizontal direction accumulator. Equation 2,3 and 4 can be used to detect the projection point as follows :

- Equation (2) determines the first projection in the vertical and horizontal direction.
- Equation (3) determines the second projection in the vertical and horizontal direction.
- Equation (4) determines Xp3 (the average value of Xp1 and Xp2) , and Yp3 (the average value of Yp1 and Yp2).

**Step 5:** Binarization of the original image by multiplying equation (1-a) * 0.6.
**Step 6:** Determine the second pupil radius approximation as shown in equation 5 and figure 4.

$$threshold = 0.6.\left\{0.52.\left[\frac{1}{size_{image}}\sum Intensity_{pixels}\right]\right\} \quad (1-b)$$

$$X_p1 = \arg\min_x \left(\sum I(x,y)\right)$$
$$Y_p1 = \arg\min_y \left(\sum I(x,y)\right) \quad (2)$$
$$X_p2 = \arg\min_x \left(\sum I(x,y)\right)$$
$$Y_p2 = \arg\min_y \left(\sum I(x,y)\right) \quad (3)$$
$$X_p3 = 0.5.(X_p1 + X_p2)$$
$$Y_p3 = 0.5.(Y_p1 + Y_p2) \quad (4)$$
$$pupil's_{radius} = \max\left(\frac{dis1}{2} + \frac{dist2}{2}\right) \quad (5)$$

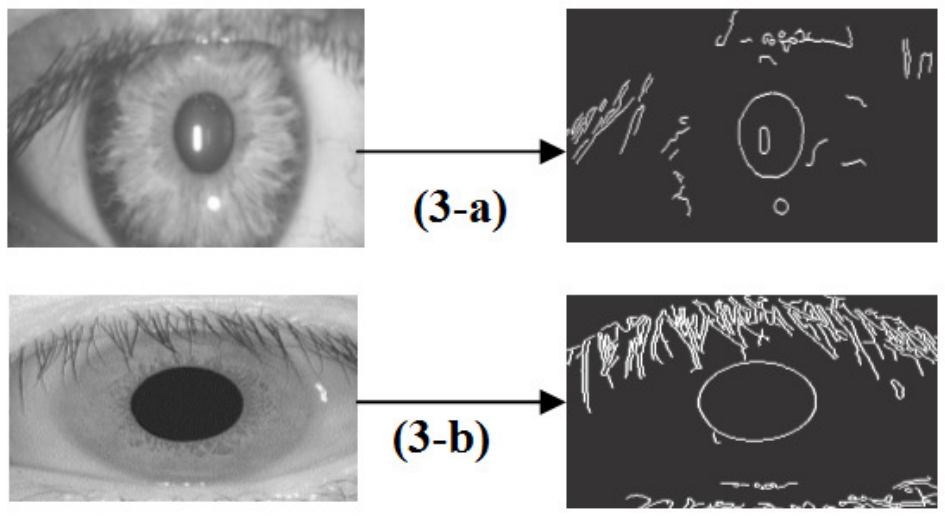

**Fig. 3:** Canny edge detection with different datasets**:** (3-a) LEI dataset, and (3-b) CASIA dataset.

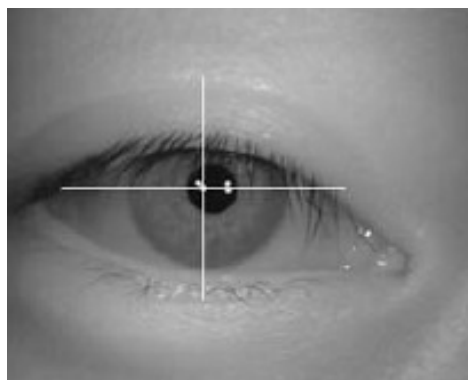

**Fig. 4:** Second pupil radius approximation.

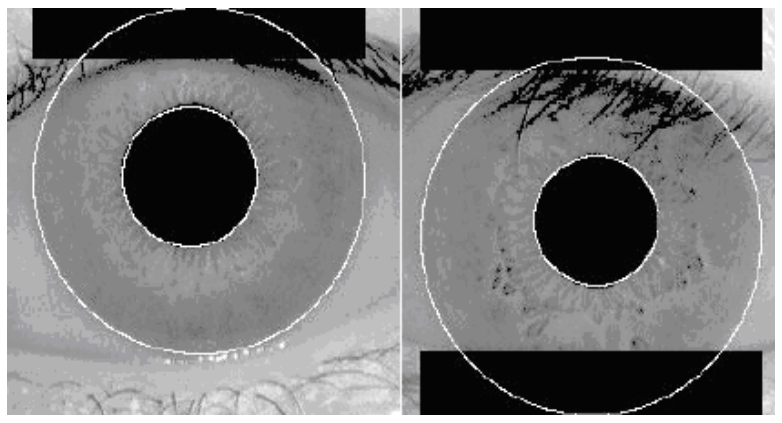

**Fig. 5:** Segmentation of various images from the CASIA database. Eyelid and eyelash regions are the black regions.

## 3.2 Iris Segmentation

The segmentation process is defined as separating the input iris image into several components so it is a very important step in preprocessing because it describe and recognize the input image. One of the best ways to separate the pupil from the whole eye is using CHT [3] in order to find the circle of the pupil . The pupil is the largest black area in the intensity image, its edges can be easily detected from the binary image by using a suitable threshold of the intensity image as shown in figure 5, 6.

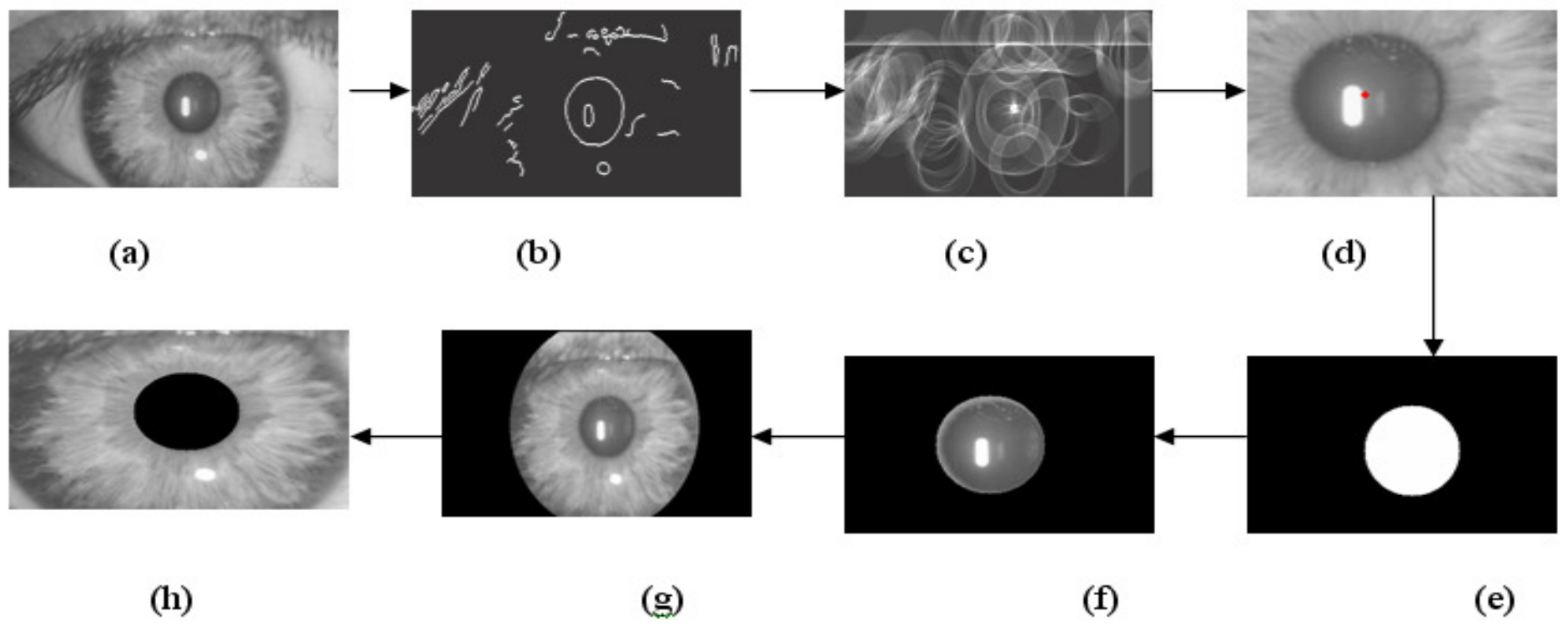

**Fig. 6:** Iris localization and segmentation stages.

<u>The steps of iris localization and segmentation are shown in figure 6 :</u>

a. The input eye image .
b. Applying Canny Edge Detection with a suitable threshold = 0.2.
c. Hough space for localization.
d. The Projection point inside the pupil of input eye image.
e. Binary matrix with 1's (white region) on the pupil region at the center and 0's (black region) on the remaining region in order to isolate the pupil.
f. The output image resulting from multiplying the input eye image with the binary matrix to detect the pupil.
g. The mask for noise detection.
h. The output image results from subtracting the input image with (f).

### 3.3 . Iris Normalization

The final step of preprocessing is normalization which transforms the iris region so that it has fixed dimensions in order to allow comparisons. Daugman's Rubber Sheet Model [17] is used for iris normalization as shown in figure 7. The rubber sheet model ramps each point within the iris region to a pair of polar coordinates (r, θ) where r is on the interval [0, 1] and θ is angle [0,2π]. The remapping of the iris region from (x, y) Cartesian coordinates to the normalized non-concentric polar representation is modeled as:

$$
\begin{aligned}
& I(x(r,\theta), y(r,\theta)) \rightarrow I(r,\theta) \\
& \textit{with} \\
& x(r,\theta) = (1-r)\, x_p(\theta) + r x_1(\theta) \\
& y(r,\theta) = (1-r)\, y_p(\theta) + r y_1(\theta)
\end{aligned}
\qquad (6)
$$

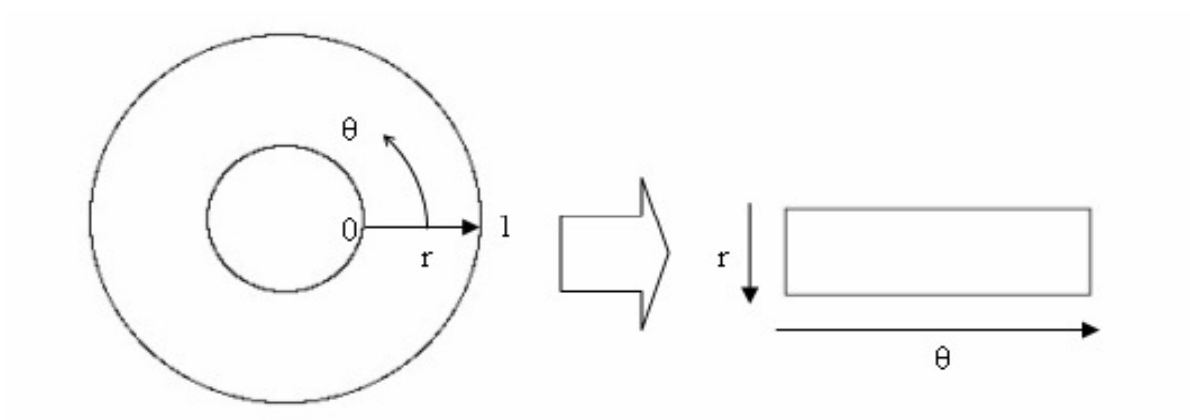

**Fig. 7:** Daugman rubber sheet model.

Where I (x , y) is the iris region image, (x , y) are the original Cartesian coordinates, (r , θ) are the corresponding normalized polar coordinates, and are the coordinates of the pupil and iris boundaries along the θ direction.

## 4. FEATURE EXTRACTION

Feature extraction is presented in order to reduce the input iris data and transfers it to feature vector. If the features extracted are carefully chosen, it is expected that the features set will extract the relevant information from the input data in order to perform the desired task using this reduced representation instead of the full size input iris image. In this paper both Local

Binary Pattern and histogram properties are presented in order to extract the features of a normalized iris image to get the feature vector required for classification using Combined LVQ Classifier.

## 4.1. Local Binary Pattern

Local Binary Pattern (LBP) is an efficient method used for feature extraction and texture classification it was first introduced by Ojala et al in 1996 [19] , this was the first article to describe LBP. The LBP operator was introduced as a complementary measure for local image contrast, and it was developed as a grayscale invariant pattern measure adding complementary information to the "amount" of texture in images. LBP is ideally suited for applications requiring fast feature extraction and texture classification . Due to its discriminative power and computational simplicity, the LBP texture operator has become a popular approach in various applications, including visual inspection, image retrieval, remote sensing, biomedical image analysis, motion analysis, environmental modeling, and outdoor scene analysis. G. Savithiri and A. Muurugan [20] presents LBP and Histogram of Oriented Gradients in order to extract the entire iris template features, they used hamming distance for matching. In this paper LBP is introduced in order to extract the iris features from the normalized iris image. The output of LBP is the feature vectors with n-dimension used as an input to combined LVQ classifier. Table 1 illustrates an example of the input sub-image with size 3×3, the center is threshold value such that, If the gray level of the neighboring pixel is higher or equal, the value is set to one, otherwise the value is set to zero. Another feature can be calculated from LBP is called contrast C where it is computed as the difference of the average pixels with threshold equal 1 and average pixels with threshold equal 0, as shown in Table 2.

**Table. 1:** An Example of LBP Computation.

| Sub-image | | | Threshold | | | Weight | | | LBP | | | | | |
|---|---|---|---|---|---|---|---|---|---|---|---|---|---|---|
| 6 | 3 | 4 | 1 | 0 | 1 | 1 | 2 | 4 | 1 | 0 | 4 | | | |
| 5 | 4 | 5 | 1 | | 1 | 8 | | 16 | 8 | | 16 | | 157 | |
| 3 | 1 | 4 | 0 | 0 | 1 | 32 | 64 | 128 | 0 | 0 | 128 | | | |

LBP = 1 + 4 + 8 + 16 + 0 + 0 + 128 = 157

**Table. 2:** An Example of C Computation.

| Sub-image | | | Threshold | | | Weight | | | LBP | | | | | |
|---|---|---|---|---|---|---|---|---|---|---|---|---|---|---|
| 6 | 3 | 4 | 1 | 0 | 1 | 1 | 2 | 4 | 6 | 3 | 4 | | | |
| 5 | 4 | 5 | 1 | | 1 | 128 | | 8 | 5 | | 5 | | 2.5 | |
| 3 | 1 | 4 | 0 | 0 | 1 | 64 | 32 | 16 | 3 | 1 | 4 | | | |

C = (( 6 + 4 + 5 + 4 + 5 ) / 5 ) – (( 3 + 1 + 3 ) / 3) = 4.8 – 2.33333 = 2.4666667

A LBP is called uniform if the transactions between "0" and "1" are less than or equal to two [21]. For example: 01110000 and 11011111 are uniform patterns**,** the patterns 00000000 (0 transitions), 01111000 (2 transitions) and 11101111 (2 transitions) are uniform whereas the patterns 110011001 (4 transitions) and 10110101 (6 transitions) are not. The histogram of the uniform patterns in the whole image is used as the feature vector [22].Uniform pattern is an extension t the original operator which can be used to reduce the length of the feature vector and implement a simple rotation-invariant descriptor.

Figure 8 shows the normalized iris image is divided into 100 smaller regions from which LBP histograms are extracted and concatenated into a single histogram.

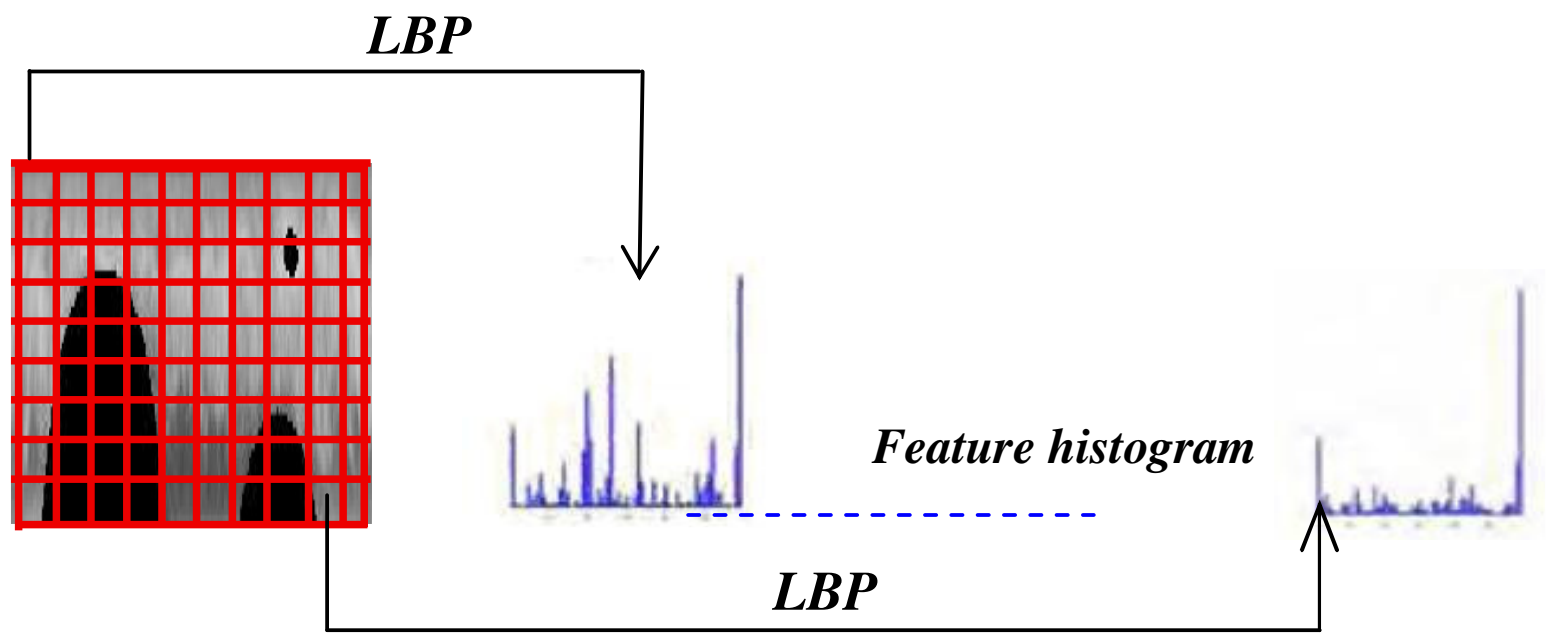

**Fig. 8:** Iris recognition based on LBP.

## 4.2. Histogram Properties

Image histogram is a first order statistics which is one pixel level, there are many statistical measures that can be extracted from the histogram using first order probability distribution [23] such as mean value among the intensity of pixel values. Histogram statistics include range, mean, geometric mean, harmonic mean, standard deviation, variance, and median. Histogram comparison statistics, such as L1 norm, L2 norm, Mallows or EMD distance, distance, Histogram intersection, Chi-square, and Normalized correlation coefficient, can also be used as texture features [23]**.** After LBP is computed, a set of the histogram features is computed together for LBP to construct the feature vector. In color texture classification with color histograms and local binary patterns, histogram techniques have proved their worth as a low cost, and level approach [24]**.** They are invariant to translation and rotation, and insensitive to the exact spatial distribution of the color pixels. These characteristics make them ideal for use in application to tonality discrimination such as the color tonality inspection using Eigen space features application [25]**.** The accuracy of histogram based methods can be enhanced by using statistics from local image [26]**.** Computation of the histogram, over the cell is needed to determine the frequency of each "number" occurring (i.e., Which pixels are smaller and which are greater than the center), and then normalize the histogram to get the feature vector .

# 5. Proposed Algorithm

The main idea of this paper is the study of how to obtain the advantages of both Statistical and Neural Network features in order to build a hybrid system depends upon the advantages of both methods. In this paper a new algorithm is proposed using LBP and histogram properties as a statistical approach for feature extraction and the use of combined LVQ classifiers as a neural network approach for classification. Once the features are extracted using both LBP and Histogram properties, an iris image is transformed into a feature vector applied to combined classifier. Figure 9 shows the structure of the proposed method, whereas the feature vector is the input to combined Learning Vector Quantization (LVQ) Classifier. LVQ is a pattern classification method, each output node is represented as a class. The weight vector of an output node is called a reference or codebook vector (CV). The LVQ attempts to adjust the weights to approximate a theoretical Bayes classifier. Input vectors are classified by assigning them as a class label of the weight vector closest to the input vector. The result obtained depends on the majority voting among several weak classifiers. The LVQ consisting of two layers; the first layer is the input layer which contains the input neurons, while the second layer is the output layer that contains output neurons. The main idea of LVQ is that for each input vector and weight vector for the n-neuron the output is calculated using the Euclidian distance to determine the winner neuron then update its weights. Figure 10 shows the steps of the proposed algorithm.

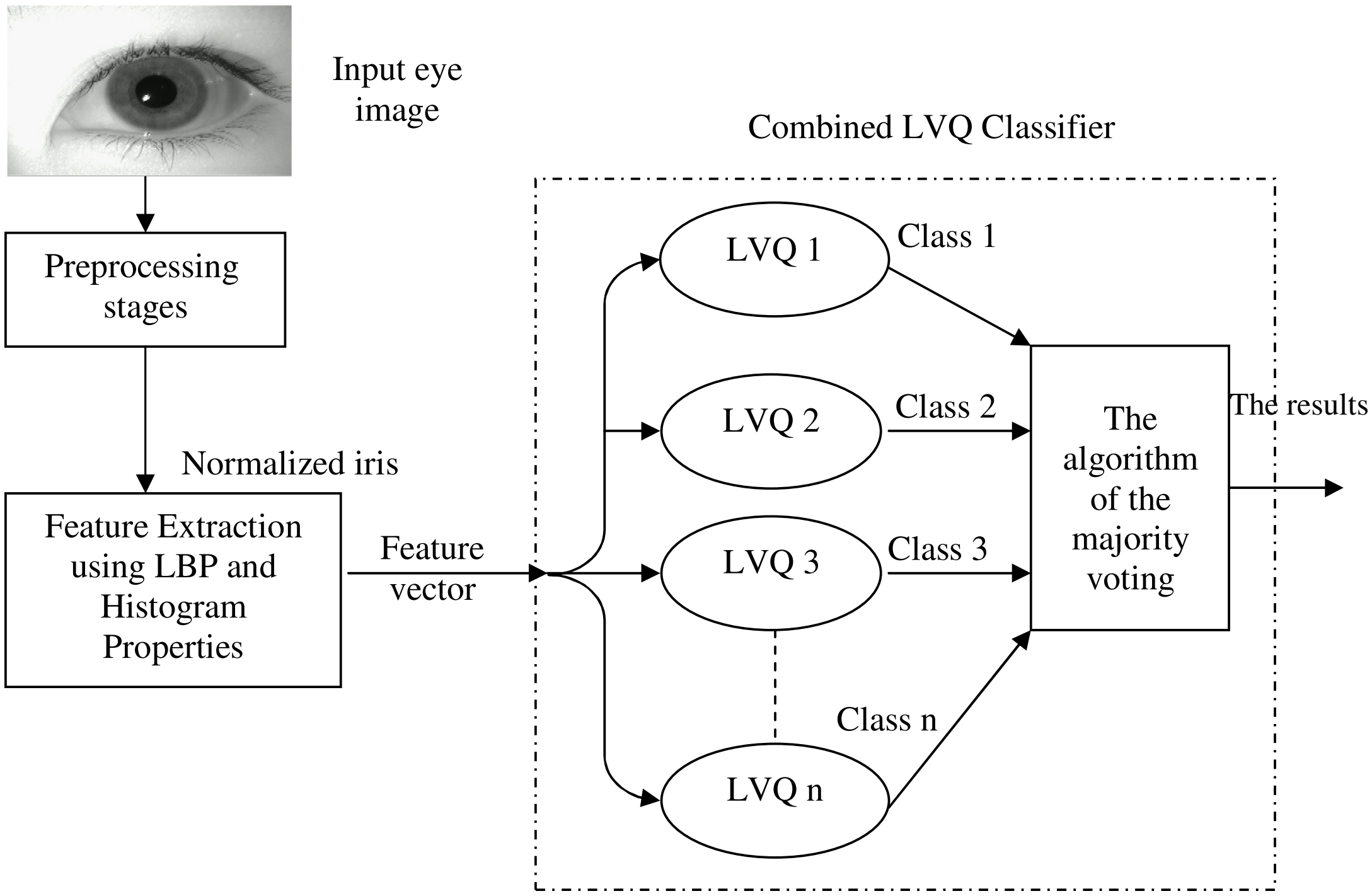

**Fig. 9:** General structure of the proposed method.

**Step 1:** Upload input iris images.
**Step 2:** The Preprocessing stages:

- Using both Canny edge detection and HCT for localization and segmentation.
- Normalization of an iris using Daugman rubber sheet model.

**Step 3:** Extract the normalized iris features using both LBP and Histogram Prosperities in order to obtain the feature vector.
**Step 4:** The Feature vectors obtained is uploaded to Combined LVQ Classifier.

- Set Nc: Number of Classifier = 3.
- Determine the MACP: Minimum Acceptable Classifier Performance.

**Step 5 :** The training and testing for each LVQ Classifier by:

- Selecting a suitable learning rate.
- Selecting a suitable NH: number of hidden neuron..
- Selecting a suitable training epoch number.
- Repeat for each LVQ Classifier (Nc=3).

**Step 6:** The output is based on majority voting among several classifiers.

- The algorithm of the majority voting

```
For m = 1 to No of image for test
    For n = 1 to No of classes
        Counter [ n ]= 0;
    End for;
        For C = 1 to No of Chosen Classifier
        K= Recognition result [ c , m ];
        Counter [ K ] = Counter [ K ] + 1;
        End for;
    Winner Class = Max ( Counter ) ;
End for.
```

**Fig 10:** The Algorithm of the Proposed Method.

## 6. Results And Discussion

Experimental results are performed in order to evaluate the performance of the proposed method. The Proposed method is implemented using MATLAB version 7.0 on Intel Pentium IV 3.2 GHZ Processor PC with 512 MB of RAM memory. Different iris datasets CASIA, MMU1, MMU2, and LEI with different extensions are used as external input iris images. The changes of LVQ classifier parameters have a high effect on the classification results. In this paper we found that the best learning rate increases the recognition rate of the system whereas the learning rate is a critical parameter that affected in the recognition process. We use a different number of learning rate (0.1, 0.2, 0.3, and 0.9) with 500 epochs and 40 hidden Neurons experiments. Table. 3 shows the results obtained with different iris images, when using (P=24, R=3), (P=16, R=2), and (P=8, R=1) and the resulting histogram runtime. The results show that the histogram properties with LBP are more reliable and effective for iris pattern description. To evaluate our proposed method, a comparative study is made:

1. Local Binary Pattern (LBP) as statistical approach for feature extraction with Learning Vector Quantization (LVQ) as Neural Network approach for classification i.e. One class of LVQ. (LBP+LVQ) .
2. Local Binary Pattern (LBP) and Histogram prosperities as a statistical approach for feature extraction with Learning Vector Quantization (LVQ) as a Neural Network approach for classification i.e. One class of LVQ. (LBP+Histogram +LVQ) .
3. Local Binary Pattern (LBP) and Histogram prosperities as statistical approach for feature extraction with Combined LVQ Classifier as Neural Network approach for classification (N=3). (LBP+Histogram + combined LVQ classifier n = 3).

Figure 11 Shows the comparison between (LBP+LVQ) , (LBP+Histogram +LVQ) . And (LBP+Histogram + combined LVQ classifier n = 3) for different input iris images and the corresponding recognition rate. We found that our proposed method has the highest recognition rate. The comparison between our proposed method and M. Shamsi et al [13] is shown in table 4 and figure 12.

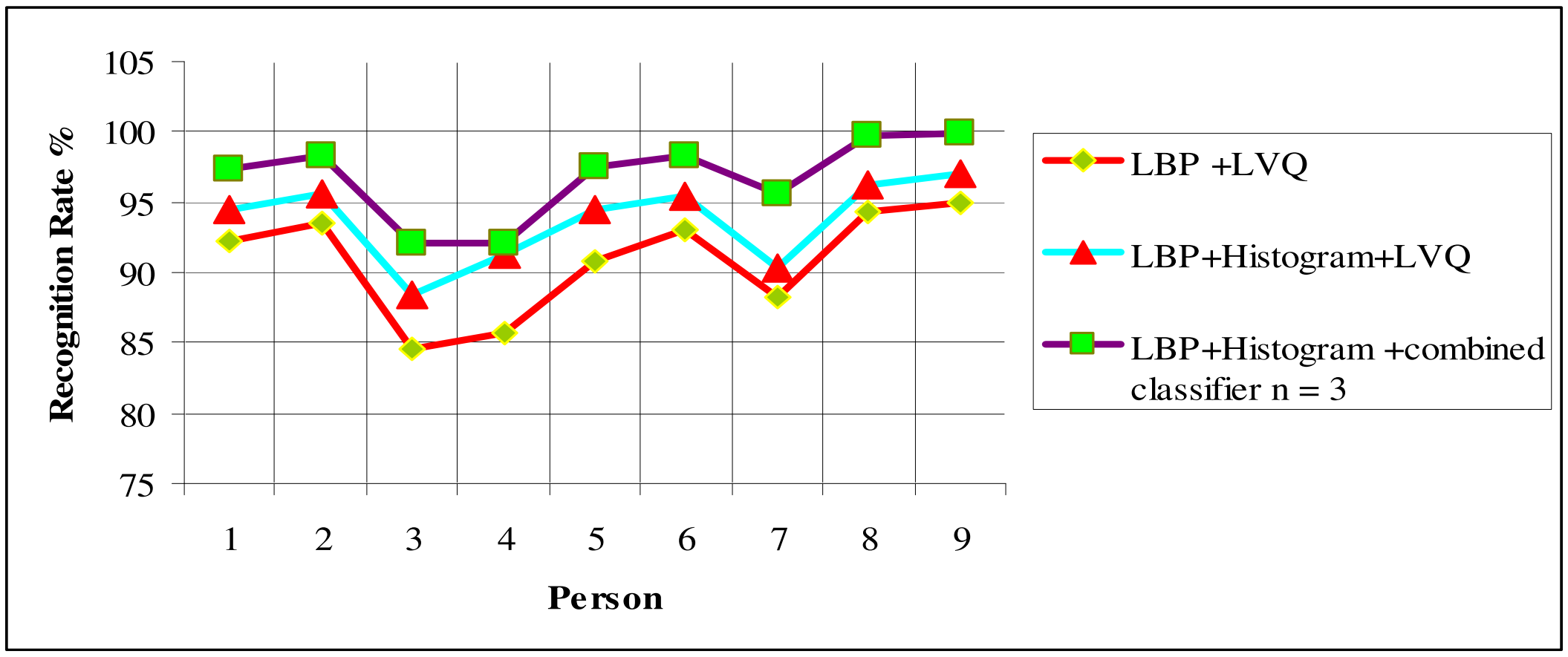

**Fig. 11:** the comparison between (LBP+LVQ) , (LBP+Histogram +LVQ) . And (LBP+Histogram + combined LVQ classifier n = 3).

Table 3: The results obtained using LBP and histogram properties.

| Iris Textures | P=24, R=3 | P=16, R=2 | P=8, R=1 | Histogram Run |
|---|---|---|---|---|
| Iris 1.JPG | 27.45 | 36.03 | 47.88 | 177.9047 |
| Iris 2.JPG | 28.32 | 37.34 | 48.23 | 178.0186 |
| Iris 3.JPG | 29.89 | 38.56 | 49.03 | 178.1479 |
| Iris 4.JPG | 30.09 | 39.44 | 51.89 | 178.1101 |
| Iris 5.JPG | 31.65 | 40.34 | 52.76 | 177.7324 |
| Iris 6.JPG | 32.56 | 41.56 | 52.99 | 179.0170 |
| Iris 7.JPG | 33.17 | 42.87 | 54.01 | 178.3433 |
| Iris 8.JPG | 34.09 | 43.87 | 55.77 | 179.0056 |
| Iris 9.JPG | 34.24 | 48.21 | 56.90 | 179.6554 |

Table 4 : The comparison between our proposed method and M. Shamsi et al [13].

| Algorithm | Recognition Rate (%) |
|---|---|
| Daugman [27] | 98.58 |
| Daugman [28] | 54.44 |
| Wildes [27] | 99.82 |
| Wildes [28] | 86.49 |
| Masek [27] | 83.92 |
| M. Shamsi [13] | 99.34 |
| Our Proposed | 99.87 |

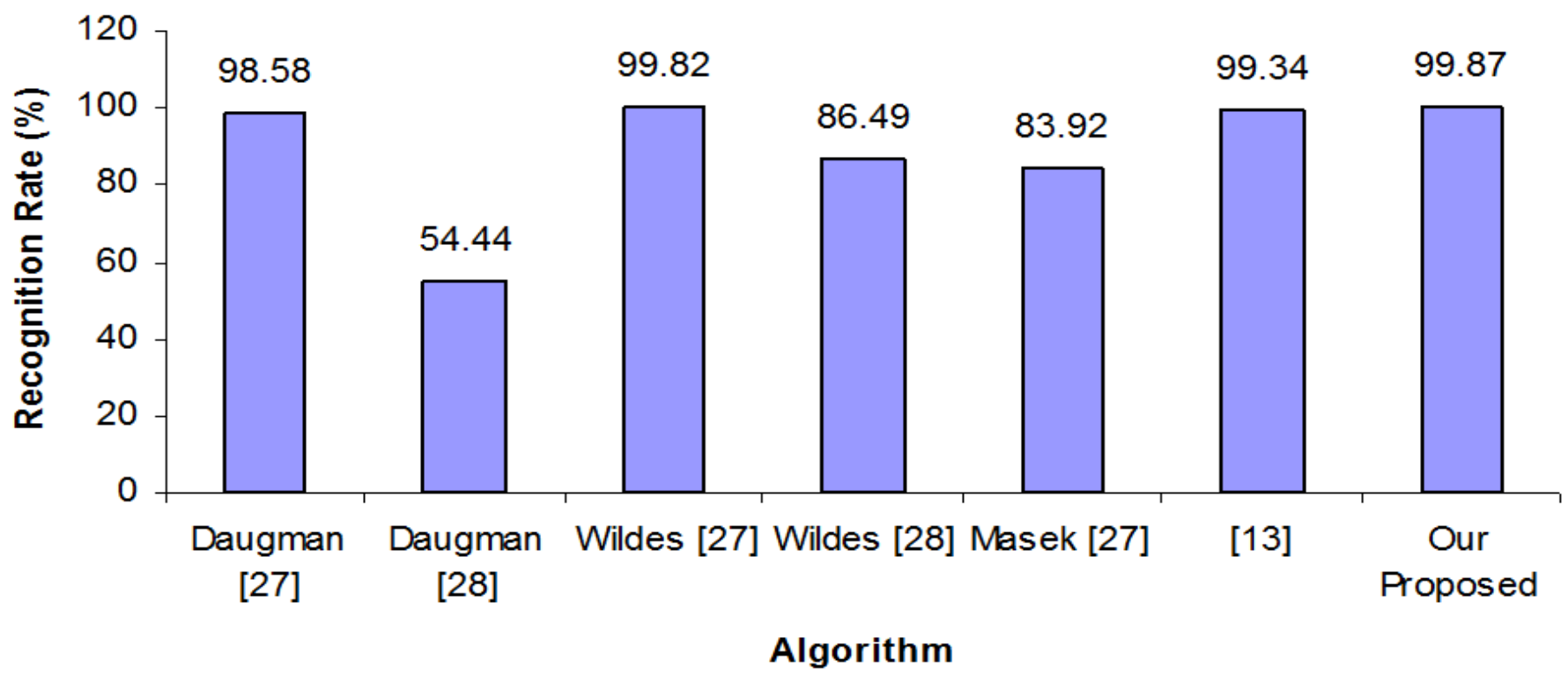

**Fig . 12**: Recognition rate of our proposed method compared with The other methods in M. Shamsi et al [13].

## 7. CONCLUSION

There are many techniques used for iris recognition differ from each others in the way of identify and verify human irises , merge two or more of these methods will produce a good result. In this paper, an efficient method for human identification and verification is presented. Furthermore, we present a system include a series of steps starting with the preprocessing step by using both Canny edge detection and Hough Circular Transform for iris localization and noise detection , then using the Local binary pattern (LBP) approach and histogram properties which actually is a very powerful feature for describing the characteristics of the iris texture image . LBP is a statistical pattern approach used to extract the iris features to get the feature vector which is the most important factor affecting the computational cost of classification .Finally, a Combined LVQ Classifier is used for classification as one of the neural network approach. The result of our proposed system is based on the majority voting among several classifiers. The comparative study shows that the proposed system has a high recognition rate compared with other methods.